# Comprehensive AI Literacy: The Case for Centering Human Agency

Sri Yash Tadimalla
College of Computing and
Informatics,
UNC Charlotte
U.S.A
stadimal@charlotte.edu

Gordon Hull
College of Humanities &
Earth and Social Sciences,
UNC Charlotte
U.S.A.
ghull@charlotte.edu

Daniel Maxwell
Cato College of Education,
UNC Charlotte
U.S.A.
dmaxwel8@charlotte.edu

Justin Cary
College of Humanities &
Earth and Social Sciences
U.S.A.
jcary1@charlotte.edu

Jordan Register
School of Professional
Studies, UNC Charlotte
U.S.A.
jtrombly@charlotte.edu

David Pugalee
Center for STEM Education,
UNC Charlotte
U.S.A.
david.pugalee@charlotte.edu

Tina Heafner
Cato College of Education,
UNC Charlotte
U.S.A.
theafner@charlotte.edu

## Abstract

The rapid assimilation of Artificial Intelligence technologies into various facets of society has created a significant educational imperative that current frameworks are failing to effectively address. We are witnessing the rise of a dangerous literacy gap, where a focus on the functional, operational skills of using AI tools is eclipsing the development of critical and ethical reasoning about them. This position paper argues for a systemic shift toward comprehensive AI literacy that centers human agency—the empowered capacity for intentional, critical, and responsible choice. This principle applies to all stakeholders in the educational ecosystem: it is the student's agency to question, create with, or consciously decide *not* to use AI based on the task; it is the teacher's agency to design learning experiences that align with instructional values, rather than ceding pedagogical control to a tool. True literacy involves teaching *about* agency itself, framing technology not as an inevitability to be adopted, but as a choice to be made. This requires a deep commitment to critical thinking and a robust understanding of epistemology. Through the AI Literacy, Fluency, and Competency frameworks described in this paper, educators and students will become agents in their own human-centric approaches to AI, providing necessary pathways to clearly articulate the intentions informing decisions and attitudes toward AI and the impact of these decisions on academic work, career, and society.

## Introduction

The fall of 2025 marks only the third anniversary since the public release of ChatGPT. Higher education is grappling with the challenges posed by the influx of generative AI technology and its impact on cognitive skills, critical thinking, perceptions of cheating, plagiarism and college writing, leading us to question the role that AI is, could, and should play in student learning. What has become abundantly clear is that AI has the potential to both enhance and diminish student learning, outcomes that are highly dependent upon the tool's alignment with instructional values and its relationship with the student's needs (e.g. the need to use or not use AI in a given context to produce a desired product or outcome or foster learning). **Comprehensive AI literacy,** building on previous efforts across various stakeholders and education levels (Chee, Ahn, and Lee, 2024; Long and Magerko, 2020; Tadimalla and Maher, 2025), an educational model that braids technical competence with ethical stewardship and the cultivation of human agency. True literacy empowers individuals not merely to use AI, but to interrogate it, ensuring technology remains a tool for human flourishing rather than a system that subverts it.

Without such a holistic approach, we risk creating a generation of

- Naive users who accept AI outputs as infallible and are unable to challenge the biases embedded within them or guide their application toward equitable ends.
- Non-users who are excluded, but not by their own empowered decisions.
- Uninformed AI champions, who reflexively advocate all uses of AI without thinking about their consequences.
- Uninformed AI opponents, who reflexively reject any and all uses of the technology without thinking about the opportunity costs.

The core problem lies in an educational approach that often treats AI as a "black box," fostering a passive and uncritical relationship with the technology. This superficial understanding neglects the complex realities of AI systems, which are not objective sources of truth but products of human choices, vast datasets, and specific socio-economic contexts (Crawford, 2021; Tadimalla and Maher, 2024). When curricula prioritize technicality over morality, they fail to address the well-documented issues of algorithmic bias, where flawed data perpetuates and even amplifies societal inequities in domains like hiring, finance, and criminal justice (O'Neil, 2016). This creates what Elish (2019) terms a "moral crumple zone," where responsibility for AI-driven failures becomes diffused, creating cognitive dissonance and leaving no one accountable. The result is an illusion of understanding, where technical proficiency is mistaken for genuine competence, leaving users vulnerable to misinformation and manipulation.

Due to the speed, breadth, and depth of these challenges, there is a growing movement in education to establish AI literacy, and in our case, one that centers on human agency. If AI is going to be everywhere, our students need to be prepared for that reality.

There is presently an all-out push to get AI into education at every level. A 2025 Microsoft report found that 93% of U.S. students have used AI at least once or twice; 42% use it weekly, and nearly a third report daily educational use. Educators reported similar use rates (2025 AI in Education, 2025). Ohio State (Helmore, 2025) has mandated that AI be in all parts of its curricula. On the heels of substantial funding from NVIDIA, the University of Florida has promised to "transform the university as a whole" with AI.[1] OpenAI is doing everything it can to make itself ubiquitous on college campuses, including partnering with Instructure to "embed AI learning experiences within Canvas," a leading learning management system used by K12 schools and institutions of higher education (Instructure, 2025; N. S. N. Singer et al., 2025). At the same time, Microsoft has pledged $4 billion on AI education in K12 (N. Singer, 2025b), while both OpenAI and Microsoft are sponsoring teacher training (N. Singer, 2025a). Other companies are more covert; Grammarly has added GenAI, blurring the line between its traditional use for grammar correction and uses that might violate academic integrity policies (Palmer, 2024). Reinforcing the pressure to integrate AI, President Trump issued an executive order promoting AI in education in late April 2025.

---

[1] https://ai.ufl.edu/about/building-an-ai-university/ (visited Aug 18, 2025).

AI Literacy, or cognate terms, is all the rage. The start of the Executive Order is a good example:

> To ensure the United States remains a global leader in this technological revolution, we must provide our Nation's youth with opportunities to cultivate the skills and understanding necessary to use and create the next generation of AI technology. By fostering AI competency, we will equip our students with the foundational knowledge and skills necessary to adapt to and thrive in an increasingly digital society. Early learning and exposure to AI concepts not only demystifies this powerful technology but also sparks curiosity and creativity, preparing students to become active and responsible participants in the workforce of the future and nurturing the next generation of American AI innovators to propel our Nation to new heights of scientific and economic achievement ("Exec. Order No. 14277," 2025)

The Order's vision of AI literacy is explicitly in the service of workforce readiness, national competitiveness, and economic advancement. Part of the purpose of the present paper is to challenge, build upon, and expand this definition.

In the meantime, universities and K12 schools are both rolling out AI literacy initiatives. Now is the time to think critically about what AI literacy should include. AI doesn't develop in a vacuum. The decisions we make now in AI education, regulation, etc., will shape what it is and does in the future. The ways we educate students about AI now will affect how they will interact with it in the future. What we need is the *right kind of AI education* to get us the *right kind of AI.*

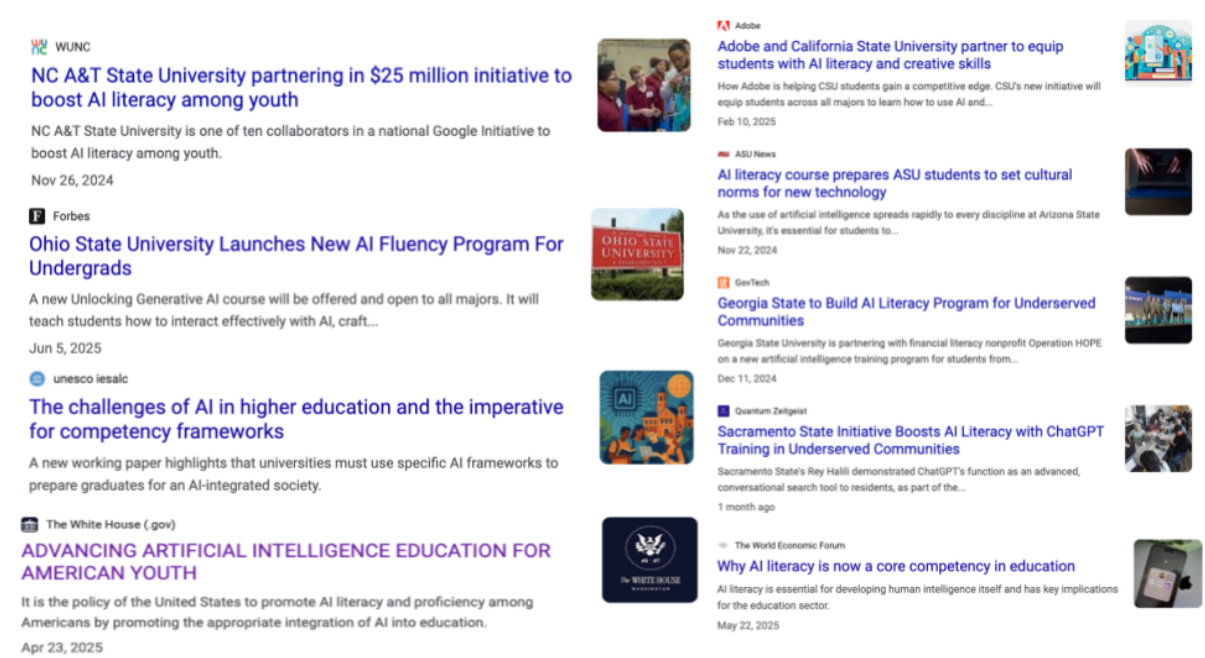

Fig 1: AI literacy Initiatives on the rise in the higher education ecosystem

The necessary solution is a robust, multi-pillar educational framework that moves beyond surface-level interaction. This requires an interdisciplinary approach that draws from the humanities and social sciences to foster deep ethical reasoning. Ultimately, the goal of AI education must be to fortify critical awareness and human agency in an increasingly automated world. The vision for a positive technological future should be explicitly human-centric, one where AI serves as a powerful collaborator to augment human intellect rather than replace it (Shneiderman, 2022). This requires teaching students to see AI as a tool to be wielded with purpose and skepticism—to know how to question its outputs, verify its claims, and recognize contexts where human judgment, empathy, and moral reasoning are irreplaceable. Therefore, we call for a fundamental reform in technology education: one that moves beyond the "how" and courageously embraces the "why." By embedding ethics and critical thinking into the core of AI curricula, we can empower a generation of citizens to be not just consumers of AI, but conscientious architects of an equitable, human-centric future.

## Background

AI is more than the datasets, algorithms, and models that are often featured in discussions about it. AI has potentially profound social implications and should be seen as a sociotechnical system, interacting with existing social and economic structures in complex ways (Selbst et al., 2019) and connecting the human and the technical throughout history.

In the following sections, we describe the socio-technical drivers that help us understand AI's positioning in society, its current impact on education, and the evolution of technology adoption in teaching and learning. Throughout these sections, we point to the need for AI literacy that centers on human agency, which we will then define for the reader.

## AI in Society

A pair of very different AI use cases can serve to illustrate the wide-ranging effects of AI. First, governments are increasingly adopting AI systems because of their efficiency and potential to deliver better services (Huq, 2020), potentially reducing the arbitrariness of regulatory enforcement and the opportunities for human corruption (Sunstein, 2022). However, AI-based decisions are notoriously opaque (Burrell, 2016), and there is a risk that AI-infused bureaucracies could become both more arbitrary and less accountable (Creel & Hellman, 2022), prioritizing efficiency over other values like compassion and individual attention (Vredenburgh, 2023). Efforts to moderate these issues can be problematic. Requiring "humans in the loop" might both be ineffective and shift blame to people who are unable to control decisions (Crootof et al., 2023; Green, 2022), while requirements that algorithms be "explainable" run into a thicket of problems caused both by the difficulty in extracting a human-understandable explanation of what the model has done, and by ethical questions about what purpose such explanation serves (Sullivan & Kasirzadeh, 2024). Moreover, and relevant here, a useful explanation of AI outputs likely requires understanding the social structures in which it is deployed (Smart & Kasirzadeh, 2024). In short, the benefits and risks of AI in governance will require not just understanding something about AI and the contexts in which it is used, but also a critical awareness that these aspects need to be understood together and as informing one another.

Second, on a daily level, the #1 use for Generative AI in 2025 is therapy or companionship (Zao-Sanders, 2025). It's easy enough to see why: chatbots are available around the clock, are relatively inexpensive, and do not judge. People also increasingly have difficulty in telling human and chatbot responses apart, and tend to rate chatbots higher (Hatch et al., 2025). At the same time, there are several lawsuits pending against LLM companies, alleging that their chatbots actively assisted deaths by suicide, including of teenagers (Wong, 2025). A recent study of LLM performance suggests that, while they tended not to respond to "high-risk" queries about suicide, their response to lower-risk queries was mixed (McBain et al., 2025). More generally, it is not clear that AI companions so much ameliorate loneliness as reproduce it online (Jacobs, 2024). These mixed results underscore the importance of understanding when and how to use - or not use - AI.

Use cases like these, and many more could be cited, make clear that AI is not simply a technical tool but a force shaping society, values, and relationships. Some of this is carefully orchestrated hype (Bareis, 2024), but many of the concerns are real. Although the details remain unclear, AI is likely to have a significant impact on everything from medical research to the workplace. Critics are broadly concerned that AI will make existing problems like healthcare disparities worse and cause widespread disruption in labor markets. AI is also expanding rapidly, with a recent report estimating that AI-associated data servers and computing will use as much energy as the entire nation of Japan by 2030 (IEA, 2025).

Definitions of AI literacy are contested, and corporate actors increasingly frame schools as a use case for generative AI. For these reasons, universities are not merely deciding how to integrate new tools into classrooms; they are determining whether future generations will inherit a democracy animated by human judgement or a polity increasingly colonized by algorithmic systems. Technology is not neutral. It erodes some cultural values, economic logics, and political interests while advancing others. This reality demands that ethical reasoning, historical consciousness, and civic purpose become the critical lenses through which AI literacy is taught. Decisions are not merely technical ones or efficiency driven; they are fundamentally civic choices.

Research across democratic theory warns that genAI (or AI) threatens the very conditions of civic trust and collective knowledge (Kay et al., 2025). Synthetic democracy, where machine-generated content saturates the public sphere with fabricated voices and consensus, erodes public accountability and representation. Misinformation and manipulation further weaken the foundations of trust in democratic institutions. In such conditions, civic education and AI literacy cannot be elective; they must function as a frontline defense against epistemic collapse of democracy's guardrails.

Collaboration with genAI should not be confused with the surrender of human agency. Yet higher education is increasingly swept up in techno-determinism, the belief that AI adoption is inevitable, and techno-solutionism, the assumption that AI will solve social and educational problems if only scaled widely. Both ideologies strip human beings of their deliberative power– a fundamental civic act. Embedding civic purposes in AI literacy requires resisting these logics, cultivating the courage to challenge automation when its use undermines dignity, equity and justice.

Just as democracy depends on systems of checks and balances to constrain power, AI governance requires civic structures that ensure accountability. Oversight councils, transparency requirements, disclosure of AI use in admission or grading, and processes for contesting automated outcomes are not bureaucratic burdens; they are civic necessities. Universities have an opportunity to model civic action by embedding accountability practices institutionally, teaching students that civic responsibility requires both agency and structural safeguards (or guardrails).

## Impacts of AI in Education

There are many examples of positive uses of AI in educational spaces, both across and within specific disciplines. General educational advancements due to AI include its ability to support differentiated instruction through personalized learning and adaptive instruction (Gligorea et al., 2023; Lata, 2024; Strielkowski et al., 2025) and its power to facilitate accessible and aligned curriculum design, instructional planning, and the development of instructional materials (NC State, 2025). It has the potential to support neurodivergent learners (Barua et al., 2022; Palumbo, 2025), multilingual access and translation (Tariq, 2024; Zaki et al., 2024), brainstorming, formative feedback, and process understanding (Hooda et al., 2022; Prompiengchai et al., 2025), while further enabling widespread creative expression to students with diverse creative abilities, facilitating the production of media and other artifacts whether in creative disciplines like arts and architecture, or as a means to demonstrate learning in varied ways (Elfa et al., 2023; Ruzieva, 2025; Wnukowska, 2024). Finally, AI has also been shown to support critical thinking and argumentation, scaffolding argumentation in ways that support critical thinking and argumentative development (Kudina et al., 2025).

With regard to disciplinary uses of AI in education, areas like social work, nursing, and counseling are leveraging AI to generate real world case studies for student learning, and to facilitate the development of critical thinking and soft skills in lieu of access to human interaction, providing students opportunities to engage in simulated, disciplinary discourse without risk to human patients (De Gagne, 2023; Glauberman et al., 2023; Singer et al., 2023). Due to its coding capabilities, AI also has the potential to support computing disciplines when used to facilitate the development of programming skills (Rajendran et al., 2023). Likewise, the mathematical disciplines are exploring the use of AI to create and analyze data sets, and support understanding of mathematical processes (Rane, 2023). Finally, AI also has the potential to offer predictive student success analytics, enabling early alert systems for instructors and administrators to target students in need of support (Cao et al., 2025).

However, as artificial intelligence tools become increasingly common in educational spaces, a number of issues arise. For example, AI tools are marketed for their ability to record and summarise meetings and other content. This recording raises concerns about surveillance, privacy and consent. Privacy is already under threat because of data analytics and ubiquitous surveillance technologies; privacy scholars are concerned that AI will make all of these problems worse (Solove, 2025). There are also issues with how AI tools integrate with existing laws. For example, if student data is used to train AI models, how does that interact with FERPA and other privacy laws? What does it mean that students can increasingly record their teachers?

The effects of AI use on learning outcomes is also unclear. There is a rise in reports suggesting that, although ChatGPT helps students write pretty essays, it does not help them learn and it risks making them dependent on the technology (Fan et al., 2025). Other research similarly suggests that students tend to cognitively offload to AI, impairing their learning (Gerlich, 2025). As one news report puts it, "the hope is that AI can improve learning through immediate feedback and personalizing instruction for each student. But these studies are showing that AI is also making it easier for students *not* to learn" (Barshay, 2025). As Ethan Mollick (Mollick, 2025) notes after surveying more such studies, AI can help or hurt in education, but we need to be intentional about prioritizing the thinking we care about.

As we see it, the value of AI tools in education must be measured by their alignment to instructional value, not solely by convenience, novelty, or application in non-educational environments. In this instance, alignment to instructional value refers to *the extent to which a tool strengthens student thinking, deepens learning, and sustains or even extends meaningful educator-student interactions*. When considered in this context, a tool's alignment with instructional values becomes inseparable from its relationship to human agency (whether understood philosophically, educationally or psychologically), of students and of instructors (Brod et al., 2023; Prieto et al., 2025). When educators thoughtfully place instructional value at the center of AI adoption, the focus intentionally shifts from *Can we use this?* to *Will this enhance learning in ways that matter?* And what does it mean to create learning experiences *that matter?* This approach positions educators as active decision-makers capable of evaluating AI tools on their merits, rather than adopting them simply because they are readily available, easily accessible, or widely promoted.

## Evolution of Technology Adoption in Learning

The history of humans working and thinking with machines is a long and complicated one. From the invention of writing, through to print, developments in mathematics and then in electronic media, the human being has always been a cyborg (Clark, 2003; Hobart & Schiffman, 1998). This means that human agency has always been at the forefront of technological innovation, as people learn to work with and through their technological systems.

Agency is also important in a learning context. Social constructivist learning theorists have posited that learning is socially constructed but also distributed across tools and symbols in that they help to mediate learning (Akpan et al., 2020; Pea, 1993; Vygotsky, 1978). Therefore, as technological advancements influence the education sector (as they always do), teaching and learning practices shift to accommodate use of the tool in the learning process.

As an example, in 1962, Bell Punch Co. introduced the first all-electronic desktop calculators, the ANITA Mark VII and Mark VIII. It was followed in 1964 by the Friden EC-130, which was solid state (using no vacuum tubes), required a fixed (not floating), decimal point, and sold for $2100 (nearly $22,000 in 2025 dollars) (Flamm, 1998). Although the ANITA calculators could only perform the basic arithmetical operations of addition, subtraction, multiplication and division, academic disciplines, industries, and consumers adapted, adjusted, and co-evolved along with this ground-breaking automation technology. Over time, developments in calculators meant neither ordinary users nor mathematics learners had to rely on a combination of arithmetical expertise, mechanical devices like the abacus and slide rule, or printed tables to do increasingly sophisticated mathematical work. By automating these practices, the calculator allowed for a range of users to engage with mathematics in novel ways.

The development of the calculator showed the risks and benefits of new technologies to human agency in education. The calculator serves as a tool to cognitively offload mathematical computations. In this way it boosts the agency of students by making room for higher order mathematical processes and thinking. At the same time, when students offload to technologies, they risk losing the skills that enable them to use those technologies effectively and ethically. In particular, there is a risk when the tools help to offload thinking essential to the learning objective. For example, if a student is developing computational skills, the calculator may impede the development of the mental models necessary for understanding and retention.

GenAI is far more capable than desktop calculators, and it is able to handle both higher order cognitive tasks and language use. This means that it will be especially important to ensure that human agency is adequately preserved as students learn. When a technological system is able to imitate or perform many of the tasks that we have traditionally associated with civic engagement and basic social interaction, the need for a literacy that goes beyond using the tool to understanding how and when to use it in ways that promote human flourishing becomes much more important.

## Differentiating AI Literacy, Fluency, and Competency

In scholarly, educational, and public discourse, the terms **AI literacy**, **AI fluency**, and **AI competency** are being used to describe escalating levels of knowledge and skill related to artificial intelligence. While sometimes used interchangeably in lay conversation, they represent a distinct conceptual hierarchy crucial for designing effective educational curricula and professional development programs.

Merriam-Webster defines literate (literacy) as "being able to read and write", and more generally as "having knowledge or competence".

**AI literacy,** building on digital literacy (Celik 2023), extends beyond technical expertise by encompassing the diverse knowledge and competencies needed to understand, interact with, and critically assess AI systems (Kandlhofer et al., 2016). This includes awareness of ethical and societal implications, such as biases, transparency, and privacy concerns that have intensified with the public availability of generative AI tools like ChatGPT (Chen 2023; Grover 2024; Touretzky et al. 2019). The technical expertise in this literacy-building approach is focused on building vocabulary and high-level understanding of the scope and technical dimensions of AI. Laying the foundation for a deeper form of knowing that one expects from AI professionals.

In contrast, fluent (or fluency) is defined as "capable of using a language easily and accurately" and "having or showing mastery of a subject or skill". Literacy is a basic ability, and fluency is the next level of skill in an area or domain.

The term **AI fluency** is ascending in prominence primarily within professional and higher education contexts, especially in fields that are not directly technological but are heavily impacted by AI. Business, law, medicine, journalism, and the creative arts are all grappling with how to integrate AI tools in a productive and ethical manner. In these domains, the goal is not to train every student to become an AI developer (competency) but to ensure that future practitioners are AI-literate users who can leverage AI to enhance their professional work through domain-specific AI fluency. Its prominence is therefore more specialized than literacy, tied to workforce development and innovation within specific industries.

**AI competency** has long been a prominent concept, but its discussion has been largely confined to highly specialized academic and industrial spheres, such as computer science and engineering departments and the research and development labs of tech companies. It is often the precursor to expertise or mastery obtained through technical AI education. The standards for competency are set by the demands of the field and are articulated in job descriptions for roles like "Machine Learning Engineer" or "AI Ethicist." While the number of people who require true AI competency is far smaller than those who need literacy or fluency, the societal reliance on this group's expertise is immense.

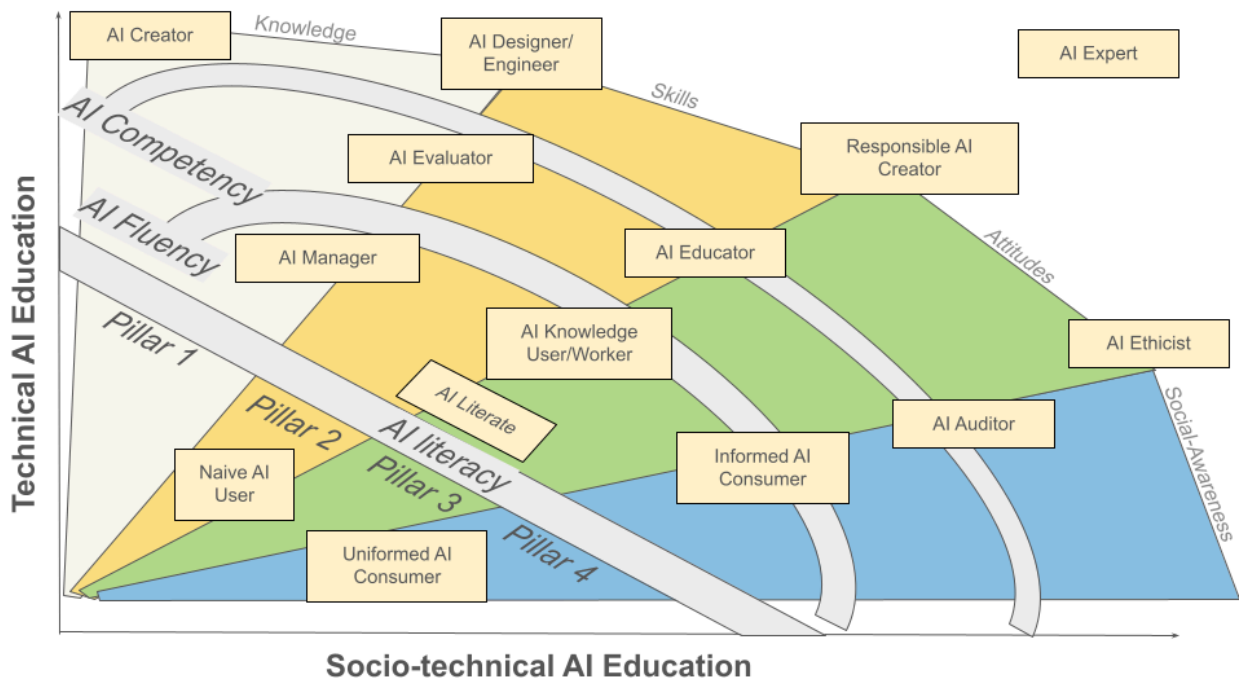

Fig 2: AI education and Carrer Pathways unlocked through AI Literacy, fluency and competency

## The 4 Pillars of Comprehensive AI Literacy

Of the aforementioned terms, AI literacy has gained the most widespread prominence in public discourse and general education (K-16). Recognizing that AI is a transformative force on par with the printing press or the internet, governments, educational bodies, and non-governmental organizations have launched numerous initiatives to promote a baseline of AI literacy for all citizens. The focus is on workforce and civic preparedness. The argument is that for the public to engage in meaningful debates about AI regulation, ethics, and deployment, a foundational, shared understanding is a prerequisite. Therefore, AI literacy has become a key objective in updating K-12 curricula and defining general education requirements in universities. Following Tadimalla and Maher (2025), we can categorize most literacy and education efforts within the 4 pillars of focus:

- Pillar 1: Understand the scope and technical dimensions of AI,
- Pillar 2: Learn how to interact with (Generative) AI technologies,
- Pillar 3: Apply principles of Critical, Ethical, and Responsible AI U.S.A.ge, and
- Pillar 4: Analyze Implications of AI on Society.

Where each pillar focuses on Knowledge mastery, skills, attitudes, perception and impact of AI. The socio-technical components of many of the initiatives mentioned above typically focus on Pillar 3 (Apply principles of Critical, Ethical, and Responsible AI U.S.A.ge), with the course subject or instructor discipline mediating the depth and scope of responsibility and accountability by individuals or organizations (Tadimalla & Maher 2024).

For a variety of reasons, many current AI literacy, fluency, and/or competency efforts minimize or even omit the fourth pillar focused on, "Analyze Implications of AI on Society," a broad, interdisciplinary topic that goes well beyond the disciplinary training of many of those in computer science departments who are tasked with teaching AI Literacy. Furthermore, AI and Society is likely to address controversial topics, where instructors may be uncomfortable leading classes and governments may have limited instructional resources. However difficult it is to support and emphasize it, we see the de-emphasis on the fourth pillar as a mistake.

## Why Pillars 3 and 4 are Critical

Tables 1 and 2 provide an overview of potential learning outcomes associated with Pillar 3 (Apply principles of Critical, Ethical, and Responsible AI U.S.A.ge) and Pillar 4 (Analyze Implications of AI on Society). We see the outcomes associated with Pillars 3 and 4 as critical for developing comprehensive AI literacy, which we expand upon, below.

| Pillar 3 | Learning Outcomes | | | | | |
|---|---|---|---|---|---|---|
| Module 9: Responsible use of AI | Responsible Use of AI | Responsible Use of AI in Education | AI Compliance | Socio-Technical frameworks | Explainable Human-Centered AI | Risks of Over-Reliance on Generative AI |
| Module 10: Security and privacy in AI | Cyber Security and AI | Physical Security and AI | AI Security and Privacy vs Secure AI | Critiquing and evaluating AI | Fairness, Accountability, and Transparency in AI | |

| Module 11: Ethical issues in AI | A Framework for Ethical Decision Making | Ethics of Data Work | AI Bias | Ethics and Values-Oriented AI Design | GenAI Case Studies: Classroom/Workplace | |
|---|---|---|---|---|---|---|
| Module 12: Integrity, Authorship and Ownership in the Age of AI | Why is Academic/Professional Integrity Important? | Code of Student/Professional Conduct | Types of Academic/Professional Misconduct using AI | Reasons and Strategies to avoid Misconduct when using AI | Understanding AI Sanctions and Policies | Intellectual Property: Authorship and Ownership |

Table 1: Learning Outcomes of Pillar 3, focused on applying principles of Critical, Ethical, and Responsible AI usage

| Pillar 4 | Learning Outcomes | | | | | |
|---|---|---|---|---|---|---|
| Module 13: Public Perception of AI | Perception and Mental Models | Public Perception of AI | Culture, Trust, and AI | Misconceptions of AI | AI and Media | AI and Identity |
| Module 14: Generative AI and the Future of Work | Shifts in AI and the Technology Industry | Digital Economy and ICT | Opportunities vs Disruptions | Un AI able: Human Vs AI | Gen AI and the Future of Work | AI in Different Sectors and Fields |
| Module 15: AI and Policy | Types of Policy and Institutions | Science Policy: Policy and Technology | History of AI Policy | Actors in AI Policy, Executive Orders | AI and Democracy | AI Regulation and Governance |
| Module 16: AI for Good, Sustainability, and Development | Sustainable Development Goals | Carbon footprint and Energy cost of Technology | AI for Social Good | Case Studies on AI for Good | | |

Table 2: Learning Outcomes of Pillar 4, focused on analyzing the Implications of AI on Society

First, only training students to use current AI models and techniques virtually guarantees that their knowledge will be obsolete by the time they graduate. For example, the *Wall Street Journal* reports that prompt engineering as a speciality is already obsolete: language models have gotten better, and companies can train existing employees to do it. As a result, there are almost no jobs that specialize in prompt engineering (Bousquette, 2025). As a recent editorial in the *Communications of the ACM* put it, "a few years from now, I expect we will look back on prompt-based interfaces to generative AI models as a fad of the early 2020s" (Morris, 2024)

Second, it is not possible to design or implement AI without considering social implications. Both those who develop AI, and those who use it, should be aware of this. Why is this? The obvious response is that the decision to shift a given task to AI will change how it is done. People drive, and they often cause accidents - up to 94% of accidents. Would replacing them with autonomous vehicles (AVs) help? This is unclear. The 94% figure, often cited by AV companies, ignores the role of things like unsafe roads, excessively large vehicles and inadequate pedestrian accommodations in causing accidents. This suggests that AVs are no panacea (Zipper, 2021). AVs will invariably introduce their own risks. Pedestrian detection systems, for example, have a harder time detecting individuals with darker skin tones (Wilson et al., 2019). Overall, current research suggests that AVs are better in some situations and worse in others (Abdel-Aty & Ding, 2024). In the meantime, there is good evidence that humans are being asked to accept legal liability for failing to intervene to override a bad decision by an AV, even when doing so would require acting much faster than they plausibly could (Crootof et al., 2023; Elish, 2019).

Although the case of AVs seems obvious, we also intend to indicate a more fundamental point. As any complex AI system is designed, those who develop it have to make a number of unforced design decisions (Biddle, 2022; Karaca, 2021). For example, consider a classifier algorithm that categorizes images of skin lesions as "low" or "high" risk for being cancerous. What factors are most important in assessing risk? It is not enough to say "trust the science," because scientific results are often ambiguous or uncertain. For the algorithm to learn what a cancerous lesion looks like, it has to be shown many examples of correctly labeled lesions. Where do these images come from, and what are the criteria for designating an image as being sufficiently high or low quality to include in the data? Meanwhile, one has to show the algorithm the right number and types of images to avoid learning the wrong classifications, as in the case of the skin cancer algorithm that learned that pictures of lesions with a ruler next to them were more likely to be cancerous.

In the meantime, other design decisions are important. For example, where does one set the threshold for becoming "high" risk? What estimated percentage chance of developing cancer puts you in the high-risk category? This decision is inseparable from a related one: the algorithm is going to be wrong at least some of the time. How does one handle that risk? Is it better to err on the side of caution, and waste money on unnecessary biopsies, or to miss some cancerous lesions? How many unnecessary biopsies trade-off with one missed lesion? Calling it a trade-off may sound uncomfortable, but it is how the system must be calibrated. This means that the system designers have to think carefully and be intentional with their design decisions, even when they are highly technical. In the meantime, algorithms are not used in a vacuum. Healthcare systems may or may not adopt them. Clinicians may or may not use them, but patient care decisions will still be affected.

AI literacy must include awareness of the design and use considerations of the field. However, if such considerations belong primarily to Pillar 4 (Analyze Implications of AI in Society), they also belong to Pillar 1 (Understand the Scope and Technical Dimension of AI), because they are tied to the technical dimensions of the system. This is what it means to say AI is a sociotechnical system. The effort to separate Pillar 4 impoverishes Pillar 1 by failing to recognize that the full scope of AI includes the societal impacts of these models. That is, any AI system worth developing will have social implications that stem directly from its technical specifications. Effective AI literacy training should not pretend that these social implications are not there, or leave students with the belief that their decisions are "purely technical." It is much better to arm students with both the tools and a sense of agency to make those decisions deliberately and responsibly.

## Centering Human Agency as a Core Component of AI Literacy

"Agency" is a complicated and multidisciplinary term, even in the context of education (Brod et al., 2023). Here we mean that efforts at technological literacy should keep the needs of humans at the center of the use of technological systems. The important part about technological systems is whether they facilitate human flourishing and are used ethically and responsibly. We should avoid treating technological development and use deterministically, as happening without human intervention.

In promoting agency, we also want to emphasize a distinction between ethical awareness—the recognition that AI is a human endeavor, the use of which has ethical implications—and ethical agency, the sense of empowered responsibility to actively shape how AI is used (Register et al., forthcoming 2026; Register et al., 2021; Register et al., 2023; Stephan et al., 2020). AI literacy needs to encompass both, moving students along a spectrum of agency towards sociopolitical understanding, civic empathy and action taking (Shor, 1993). Of course, within a given context, individuals may exercise their agency differently, and according to their goals or expectations related to that context. That is, a person might recognize the ethical issue, but may choose to take action in a way that impacts the system or an isolated incident, or they may choose not to act at all because they do not feel that they have the responsibility to do so based on their role, or some other confounding factor. Importantly, a core element of AI literacy is being able to navigate these situations intentionally with consideration for the impacts on the self, others, and society more broadly. This frames AI agency as a critical dimension of AI literacy—one that, in educational contexts, is closely linked to the alignment with instructional value and to broader commitments to civic engagement and responsibility.

## Focus on Agency to Maximize Instructional Value

Evaluating AI through a lens of alignment with instructional value preserves human agency and enables educators to balance both the benefits and limitations of these tools without assuming an outright pro- or anti-technology stance. For instance, in a humanities course, translation tools may be used to enable access to complex texts for multilingual learners, but their instructional value depends on whether students still deeply engage with the cultural and historical context of those texts. Similarly, AI-assisted research may help a user locate relevant sources, yet the core instructional value emerges when students are guided to further evaluate credibility, relevance, or evidence of bias within AI-recommended sources. AI uses that bypass human inquiry processes, short-circuit critical thinking, or provide unclear, unverifiable results diminish both instructional value and human agency. By grounding AI literacy efforts within a framework of instructional value, educators can ensure that societal and civic implications are not treated as an optional add-on, but as a core criterion for determining whether and to what extent AI truly serves the purposes of teaching and learning.

Sometimes, AI will not be the right tool, either in education or the workplace. This highlights a key component of agency: the decision to use or not use a given tool. In a now-classic paper, Sally Wyatt (Wyatt, 2003) raised the question of the Internet Non-User and their importance to both the design of the Internet and campaigns around it (which included a massive push to get it into schools). A later paper by Christine Satchell and Paul Dourish (Satchell & Dourish, 2009) points out that understanding non-users is important for HCI (Human-Computer Interaction) as a branch of computer science. Both papers emphasize that non-users are typically conceived as lagging adopters of a technology - future users who need to be nudged into adoption.

A similar narrative has taken hold around AI literacy. This is evident in industry financing for AI literacy initiatives and in federal policy like the recent Executive Orders. This narrative then drives education initiatives for which everyone in all disciplines is assumed to be going to use AI and needs, primarily, to learn to use it better. This narrative is a mistake that hobbles true AI literacy. Everyone will live in a world affected by AI, but not everyone will, or should, use AI. This means that AI literacy initiatives need to respect the agency of two kinds of non-users. It is far too early to know for sure what AI use does in education, but the early data is concerning. One recent MIT in-between study had students write essays under three conditions: using an LLM, using a Google search, and a no-tech group. Brain imaging showed that the AI group demonstrated decreased brain activity when presented with cognitive tasks compared to their peers in the other groups (Kosmyna et al., 2025).

In such an environment, two things are clear. First, it is far too early for university-wide, top-down imposition of AI into all learning. Second, AI literacy initiatives need to be critically aware of the role of AI in education and learning processes. The focus must be on student education and learning outcomes. These goals should drive how and whether AI is used, and not the other way around. Recent empirical evidence shows that not using AI is an important part of workflows, even among computing professionals who are receptive to AI or use it in other contexts (Cha & Wong, 2025). As Cha and Wong note, a variety of factors, ranging from system and design process specifications to professional responsibilities to larger social, legal and regulatory factors influence decisions not to use AI. Teaching students how to work through these social and contextual factors is an essential component for AI literacy, even if one's only goal is to prepare them for the workforce.

Respecting the agency of AI non-users is also important for system design. As Cha and Wong put it, by addressing what AI should not do, designers and researchers can prioritize the preservation of human agency, professional judgment, and ethical responsibilities. In other words, those who are involved with using AI and developing AI (the very people at the core of AI literacy initiatives) need to be taught to think about when AI is inappropriate. This attention will make better, more human-centered AI.

Some people actively resist new technologies like AI, for which they are often portrayed as anti-technology Luddites. As Satchell and Dourish note, this common framing misreads history. The Luddites were not opposed to technology - they were opposed to exploitative labor practices. As a current paper on technology refusers puts it, "When refusers think strategically about the actions they need to take to bring about better technological futures, they are engaging in a design process that is pragmatically oriented and considers alternatives" (Zong & Matias, 2024, p. 2). Respecting non-users is part of the civic education component of AI literacy, and is accordingly essential for any AI literacy initiatives that are relevant to the world that we want our students to help create.

AI Literacy initiatives need to be designed to respect the AI non-user, because that is a real person and not just a blip on the way to universal AI adoption. This is not just about understanding the social context of AI use and the AI industry (though it is that). It's about understanding that a healthy part of AI literacy is understanding why people might not use it, and what we can learn from them. AI Competency has to include knowing how and when not to use AI.

## Treat AI Ethics as a part of Civic Education

Because AI systems are complex, sociotechnical artifacts, the ethics required is about social relations. AI systems are trained on socially-produced data - mostly from the Internet - and data relations are themselves inherently social because what a system learns about one person can be used to make inferences about others (Viljoen, 2021). So too, as AI systems are implemented more broadly, they will touch all aspects of social life. This is why we led with the examples of governance and therapy chatbots - they exemplify the need to use a sociotechnical lens to understand the ethics of AI.

This is not to take sides in a normative debate about what kind of ethical theory (utilitarianism or deontology, etc.) is appropriate for AI systems; rather, it is to point to a general approach. There is already substantial literature exploring these issues, which should be central to AI education. Some of them arise as systems are developed. For example, there is evidence that language models perform less accurately on languages that are structurally unlike English (Helm et al., 2024). Image generation algorithms tend to default to cultural stereotypes about South Asia (Qadri et al., 2024). These are problems that arise in the development of models and depend on questions of training data.

Sometimes the ethical issues arise when systems are deployed and interact with existing social problems. For example, reliance on algorithms to determine opioid prescribing has the laudatory goal of reducing opioid dependence and deaths, but also can interfere with doctor-patient relationships and deny pain medicine to those who desperately need it (Pozzi, 2023). A hospital triaging algorithm designed to provide better care for patients with complex health needs ended up failing to notice the poor health of Black patients because its method of identifying illness picked up on the ways Black patients had been underserved by the healthcare system in the past (Obermeyer et al., 2019).

Most infamously, facial recognition software is less accurate on darker-skinned people (Buolamwini & Gebru, 2018), and there have been repeated media reports of people of color wrongfully arrested and detained based on incorrect facial recognition (MacMillan et al., 2025). This is one of many ways that data-driven policing can disproportionately hurt minoritized populations (Selbst, 2017), as the data that feeds algorithmic systems reflects the carceral system's long history of racial bias (Mayson, 2019).

In short, the ethical component of AI literacy must be outward facing, looking squarely at the social benefits and risks of using AI. Students learn about AI when they learn about what happens when it's used in the real world.

The same dynamics that destabilize civic trust and redistribute power in society at large also risk reshaping education. Civic education research reveals stark inequities in civic empowerment opportunities (Kawashima-Ginsberg & Sullivan, 2017; Levinson, 2010, 2012; Rubin & Hayes, 2010). The gaps now threaten to shape AI literacy, replicating in educational settings the very inequalities AI perpetuates in broader civic life. Elite institutions experiment with AI as innovation, while historically marginalized communities are more often subject to AI as surveillance, grading, or automated control (Duberry, 2022). The result is a civic empowerment gap: whereas traditional approaches may foster institutional efficacy for white students, they tend to cultivate protest and voice among Black and Brown students (Bañales et al., 2020; Nelson, 2019). Without justice-oriented civics (Westheimer & Kahne, 2004), AI adoption risks exacerbating these disparities rather than addressing them. Conversely, when civic-mindedness guides AI literacy, AI can serve as a tool of civic action to empower vulnerable communities, such as in participatory uses of AI for climate modeling, if and only if these communities' voices are centered in the design and governance of AI systems (Sieber et al., 2025).

Embedding civics in AI literacy requires building pedagogies that affirm human experience, not displace it. Only such a strategy will safeguard human agency. For example, storytelling, listening, and empathy are essential counters to synthetic text and fabricated content. Verification, informed skepticism, and epistemic vigilance, what two of us have called the CIVIC pillars (Heafner & Maxwell, in press), become a framework for democratic habits of mind. Students should be taught to ask not only if this is a credible source, but also who created the source, whether it is human or synthetic, and to what end. Critical inquiry reframes AI literacy away from prompt optimization and toward civic inquiry: Should this task be automated? Whose values are embedded in this system? What are the democratic costs of efficiency?

The civics of AI literacy also requires grappling with the role of government in shaping technological futures. Government agencies are already deploying AI at scale (U.S. Government Accountability Office, 2025), raising questions of oversight, privacy, and public accountability (Duberry, 2022). Internationally, deliberations over AI's role in society stress global equity and human rights (Sieber et al., 2025). Higher education must prepare students to critically question: What responsibilities should the government assume in regulating AI? What happens when corporate innovation races ahead of democratic oversight? Civics offers a corrective by training students in inquiry, deliberation, and structural critique. Embedding civic responsibility means empowering future leaders to interrogate and contest the institutional and corporate structures that shape AI's development as well as envision participatory models where citizens have a voice in its design and implementation within their communities (Arana-Catania et al., 2021).

AI unsettles the foundations of trust and redistributes power. Deepfakes and synthetic media corrode trust in evidence, blurring the lines between fact and fabrication (Fallis, 2021; Kaplan, 2020). Algorithmic decision-making undermines confidence in fairness, while the consolidation of data infrastructures in the hands of a few corporate, private actors centralizes power and limits accountability (Cohen, 2019, 2025).

Scholars warn of the rise of synthetic opinion, in which AI-generated voices manufacture consensus, simulate legitimacy, and displace authentic civic participation. Such dynamics not only threaten democratic practice but risk producing a synthetic polity that erodes public trust (Calvo & Saura García, 2024; Saura García, 2024). Civics within AI literacy must prepare students to analyze who benefits from AI, who is harmed, and what kinds of discerning trust, if any, should be extended to synthetic systems.

Civic responsibility is not constrained by national borders. AI literacy includes global civic orientation, attuned to inequities of access, digital colonialism, and planetary sustainability. Civics emphasizes human rights, inclusivity, and global equity, teaching students to see AI as a planetary issue: one that shapes energy consumption, climate futures, global governance, and international governance.

Embedding civic education and social responsibility in AI literacy is not enrichment; it is survival. Without it, universities risk producing graduates fluent in prompt engineering but illiterate in democracy, adept at leveraging synthetic discourse but unable to discern human truth. The central concern is not whether or not students will use AI. The question is whether they will have the civic, ethical, empathic, and human capacities to decide when its use undermines the very conditions of civil and democratic life. Civics, rooted in responsibility, justice, and human dignity, can be the compass guiding higher education's engagement with AI.

## Conclusion

At the heart of our proposed framing of AI Literacy is human agency—the empowered capacity for intentional, critical, and responsible choice. Students must be equipped to interrogate the very nature of knowledge in an age of synthetic text, asking not just "Is this answer correct?" but "Where did this information come from? What biases are embedded within it? And how does it change how we know what we know?" AI Literacy initiatives that promise to infuse AI adoption and U.S.A.ge in the entirety of curricula, or into all parts of a university education, take agency away from individual educators and students. The content of AI literacy must emphasize that AI will not always be an appropriate tool. This is a basic component of workplace readiness. For a variety of reasons, many current AI literacy, fluency, and/or competency efforts minimize or even omit the fourth pillar, "Analyze Implications of AI on Society" a broad, interdisciplinary topic that goes well beyond the disciplinary training of many of those in computer science departments who are tasked with teaching AI Literacy. Furthermore, analyzing the Implications of AI on society is likely to address controversial topics of the time, which may make instructors uncomfortable or unprepared. However difficult it is to engage faculty to have these conversations, we see the de-emphasis of the fourth pillar as a mistake. Because AI is a sociotechnical system, individual agency must be understood within a broader societal context. Therefore, comprehensive AI literacy is inseparable from civic education. It demands a direct confrontation with the complex interactions between AI, capitalism, and democracy. Students must analyze how these systems are deployed in governance, their impact on public discourse, and the ethical ramifications of using public data—often without explicit public consent—to train proprietary models. This civic lens provides the necessary framework for students to engage with the challenges of regulation, safety, and governance. The popular concept of the "AI Alignment" problem—ensuring AI aligns with human values—is fundamentally a civic question. A robust civic education empowers future generations to participate in the democratic process of defining what those social values are and how technology should serve them. This is grounded in a deep sense of ethics and a recognition of our shared humanity, ensuring that the pursuit of technological advancement does not come at the cost of equity, dignity, and justice. AI literacy efforts within education must be designed to respect educators and students, who must be able to choose when, to what extent, and when not to use AI, through intentional consideration of alignment with learning outcomes or other educational goals. Educators and students need to be empowered to make this decision through dialogue.